\documentclass[twoside]{article}
\usepackage[accepted]{aistats2018}
\usepackage{booktabs} 
\usepackage{mathtools,bm,dsfont,amsfonts,amssymb,accents}
\usepackage{graphicx}
\usepackage{color}
\usepackage{url}
\usepackage{epstopdf}
\usepackage{multirow,bigstrut,rotating}
\usepackage{algorithm}
\usepackage[noend]{algorithmic}
\usepackage{amsthm,theoremref}
\usepackage{footnote,refcount}
\makesavenoteenv{tabular}
\makesavenoteenv{table}
\usepackage{capt-of,caption} 
\usepackage{subfiles}

\newcommand{\darkgray}[1]{{\leavevmode\color[RGB]{140,140,140}{#1}}}

\long\def\comment#1{}

\newfloat{algorithm}{t}{lop}
\newcommand{\CMNT}[1]{\hfill \darkgray{\(\triangleright\) {\small{{#1}}}}}
\let\oldFootnote\footnote
\newcommand\nextToken\relax
\renewcommand{\eqref}[1]{Eq.~(\ref{#1})}

\renewcommand\footnote[1]{%
	\oldFootnote{#1}\futurelet\nextToken\isFootnote}
\newcommand\isFootnote{%
	\ifx\footnote\nextToken\textsuperscript{,}\fi}

\newtheorem{lemma}{Lemma}
\newtheorem{prop}{Proposition}
\DeclareMathOperator*{\argmin}{argmin}
\DeclareMathOperator*{\argmax}{argmax}

\newcommand{\expect}[2]{\mathbb{E}_{#1}\left[{#2}\right]}
\newcommand{\cov}{\text{cov}} 
\newcommand{\covargs}[2]{\cov_{#1}\left[{#2}\right]}

\newcommand{\R}{\mathbb{R}}
\newcommand{\N}{\mathbb{N}}
\newcommand{\one}{{\mathds{1}}}
\newcommand{\1}[1]{{\one_{\{{#1}\}}}}
\newcommand{\prob}{\mathbb{P}}
\newcommand{\probarg}[1]{\prob\left[{#1}\right]}

\newcommand{\var}[1]{{\text{\ttfamily#1}}}
\newcommand{\method}[1]{{\textsc{#1}}} 

\DeclareMathSymbol{\widetildesym}{\mathord}{largesymbols}{"65}
\newcommand\lowerwidetildesym{%
	\text{\smash{\raisebox{-1.3ex}{%
				$\widetildesym$}}}}
\renewcommand\widetilde[1]{%
	\mathchoice
	{\accentset{\displaystyle\lowerwidetildesym}{#1}}
	{\accentset{\textstyle\lowerwidetildesym}{#1}}
	{\accentset{\scriptstyle\lowerwidetildesym}{#1}}
	{\accentset{\scriptscriptstyle\lowerwidetildesym}{#1}}
}

\newcommand{\seed}{{S}}
\newcommand{\unlbld}{{U}}
\newcommand{\parent}{{\rho}}
\newcommand{\parmat}{{T}} 
\newcommand{\expparmat}{{\bm{\parmat}}} 
\newcommand{\conf}{{\sigma}} 
\newcommand{\dist}{{D}} 
\newcommand{\f}{{f}} 
\newcommand{\fhat}{{\hat{\f}}} 

\newcommand{\yhat}{{\hat{y}}}
\newcommand{\nclasses}{{L}}
\newcommand{\bias}{{b}} 
\newcommand{\Bias}{{\bm{b}}}

\newcommand{\lap}{{\mathcal{L}}} 
\newcommand{\nei}[1]{{Nei({#1})}}
\newcommand{\pnlty}{{q}}
\newcommand{\prior}{{\bm{\rho}}}
\newcommand{\Y}{{\mathcal{Y}}}
\newcommand{\ancestor}{{\alpha}}
\newcommand{\Ancestor}{{\mathcal{A}}}
\newcommand{\dummy}{{r}}
\newcommand{\nullstate}{{\varnothing}}
\newcommand{\rv}{{Y}}
\newcommand{\lp}{{\textsc{lp}}}
\newcommand{\ip}{{\vphantom{}}} 
\newcommand{\secref}[1]{Sec.~\ref{#1}}
\newcommand{\Atilde}{{\widetilde{A}}}

\newcommand{\tblsmall}{\fontsize{6pt}{7pt}\selectfont}
\newcommand{\tbltiny}[1]{\fontsize{5pt}{7pt}\selectfont{#1}}
\newcommand{\tblsmallb}{\fontsize{7.5pt}{8pt}\selectfont}
\newcommand{\tbltinyb}[1]{\fontsize{6.5pt}{8pt}\selectfont{#1}}

\begin{document}

%

%

\twocolumn[

\aistatstitle{Semi-Supervised Learning with Competitive Infection Models}

\aistatsauthor{ Nir Rosenfeld \And Amir Globerson}

\aistatsaddress{ Harvard University \And Tel Aviv University } ]



\begin{abstract}
The goal in semi-supervised learning is to
effectively combine labeled and unlabeled data.
One way to do this is by encouraging smoothness
across edges in a graph 
whose nodes correspond to input examples.
In many graph-based methods,
labels can be thought of as propagating
over the graph,
where the underlying propagation mechanism 
is based on random walks or on averaging dynamics.
While theoretically elegant,
these dynamics suffer from several drawbacks which can hurt 
predictive performance.

Our goal in this work is to explore alternative mechanisms for propagating labels.
In particular, we propose a method based on  \emph{dynamic infection processes},
where unlabeled nodes can be ``infected'' with the label of their already infected neighbors.
Our algorithm is efficient and scalable,
and an analysis of the underlying optimization objective reveals a 
surprising relation to other Laplacian approaches.
We conclude with a thorough set of experiments
across multiple benchmarks and various learning settings.

\end{abstract}


\section{Introduction}
The supervised learning framework underlies much of the empirical success
of machine learning systems.
Nonetheless, results in unsupervised learning have demonstrated
that there is much to be gained from unlabeled data as well.
This has prompted considerable interest in the 
{\em semi-supervised learning} \cite{chapelle2006semi} setting,
where the data includes both labeled and unlabeled examples. 
Methods for semi-supervised learning (SSL) are especially useful for applications in which
unlabeled examples are ample, but labeled examples are scarce or expensive. 

One of the most wide-spread approaches to SSL, and our focus in this paper, is the class of {\em graph-based methods}. In these, part of the problem input is a graph that specifies which input points should be considered  {\em close}.
Graph-based methods assume that proximity in the graph implies similarity in labels. There are many variations on this idea \cite{belkin2004semi,blum2001learning,rifai2011manifold,sindhwani2005beyond}, each using smoothness and graph distance differently. However, they all share the intuition that  the classification function should be smooth with respect to the graph.

One way for encouraging smoothness 
is by optimizing an objective based on the graph Laplacian.
This is prevalent in classic SSL methods such as Label Propagation
(LabelProp) \cite{labelprop} and its variants \cite{llgc,adsorption,mad},
as well as in recent deep graph embedding methods \cite{planetoid,deepwalk}.
In some cases, the Laplacian objective can be interpreted
as the probability that a random walk terminates at a certain state.
In others, the objective can be expressed as a quadratic form
which can be optimized by iterative local averaging of labels.
The optimization process can hence be thought of as propagating labels
under a certain averaging dynamic process,
whose steady state corresponds to the optimum.
Due to their elegance, computational properties, and empirical power,
random walks and local averaging have become 
the standard mechanisms for propagating information in many applications.
Nonetheless, they have several shortcomings, which we address here.

\begin{figure*}[t]
\begin{center}
\includegraphics[width=\textwidth]{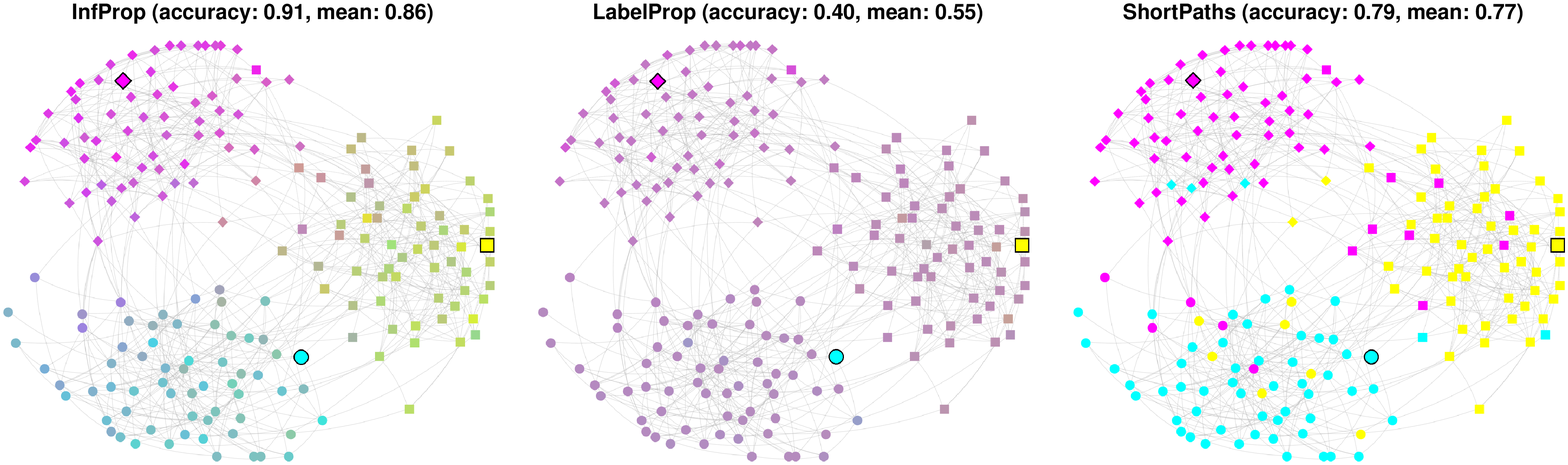}
\caption{For graphs with weakly inter-connected components,
infection dynamics (our method, left) propagate labels better
than random walks (middle) or shortest paths (right).
Labeled nodes are outlined,
shapes denote true labels,
and probabilistic predictions are encoded by CMY color values.
See supp. material for more details.}
\label{fig:illustrative}
\end{center}
\end{figure*}


First,
many of the guarantees of such methods
hold only for undirected graphs.
For directed graphs, the Laplacian is not necessarily PSD,
meaning that the objective is no longer convex,
and that the quadratic smoothness interpretation breaks down.
Optimization in directed graph Laplacians is much harder and far less understood \cite{vishnoi2013Lxb},
and sampling is computationally prohibitive,
slow to converge, and unstable \cite{lin2008multirank,lofgren2014fast}.

Second,
such methods were originally designed for graphs that approximate the density of the data in feature space.
As such, they can fail when applied to real graphs,
especially large networks with a community structure.
This is because random walks are prone to get stuck in local neighborhoods \cite{broder1994trading},
because visiting all nodes can require an expected $O(n^3)$ steps \cite{aleliunas1979random},
and since the limit distribution can be uninformative
for large graphs \cite{vonLuxburg2010getting}
or when labels are rare \cite{nadler2009semi}.

Third,
extending such methods beyond the vanilla multiclass setting
has proven to be quite challenging.
For instance, outputting confidence in predictions is possible,
but leads to extremely low values \cite{mad}.
Label priors can be utilized, but only 
in determining the classification rule,
rather than being incorporated into the model \cite{labelprop}.
Most methods for an active semi-supervised setting
are either heuristic \cite{guillory2009label}
or based on pessimistic worst-case objectives \cite{guillory2012active}.
Finally, supporting structured labels is far from straightforward and can be computationally demanding \cite{altun2006maximum}. 

Due to the above,
in this work we advocate for considering alternative mechanisms for propagating labels over a graph, and propose an approach
which addresses most of the above issues. 
Our method,
called \emph{Infection Propagation} (InfProp),
views the process of labeling the graph as a \emph{dynamic infection process}.
Initially, only the labeled nodes are ``infected'' with their known labels.
As time unfolds, infected nodes can, with some probability, infect their unlabeled neighbors.
When this happens, the unlabeled nodes inherit the label of their infector.
Labeled nodes can therefore be viewed as competing over
infecting the unlabeled nodes with their labels.
Since the infection process is stochastic, we can calculate the probability that a given node was infected by a given label,
and label the node according to this probability. 

InfProp is motivated by the idea that
different graph types may require different dynamics for efficient propagation of information.
It is inspired by propagation dynamics
found in the natural and social worlds,
and draws on the successful application of infection models
in different contexts
\cite{kempe2003maximizing,mgr2010inferring,chen2010scalable}.
InfProp is especially efficient for graphs with 
highly intra-connected but lightly inter-connected components,
a characteristic of many real-world networks.
Fig.~\ref{fig:illustrative} illustrates this
for a small synthetic random network with three clusters
(see  supplementary material for details).
As can be seen, InfProp propagates information correctly, even
when the seed set is very small.
In comparison,
LabelProp provides uninformative and almost uniform predictions
which are prone to error,
and shortest paths over the weighted graph err due to cross-cluster links.

InfProp uses infection probabilities for labeling;
these, however, turn out to be \#P hard to compute exactly.
We therefore provide a fully polynomial-time randomized approximation scheme (FPRAS).
Our solution exploits an equivalence 
between the infection process and shortest paths in random graphs.
The resulting algorithm is easy to parallelize, making the method highly scalable.
It also extends to various learning settings, such as multilabel prediction and active SSL.

In \secref{sec:related} we analyze the optimization objective
underlying the propagation of labels via infection dynamics,
highlighting an intriguing connection graph Laplacians.
Our analysis shows that InfProp can be viewed as optimizing a quadratic objective,
in which weights are seed-specific and related in an intricate manner to the underlying diffusion process.
We conclude with an extensive set of experiments in multiple learning settings
which demonstrate the effectiveness of our approach.
\section{Propagating Labels with Infections} \label{sec:model}
In this section we present our infection-based method
for semi-supervised learning.
We are given as input
a directed weighted graph $G=(V,E,W)$,
as well as a subset of labeled nodes
$\seed \subseteq V$ referred to as the \emph{seed set}.
Each seed node $s \in \seed$ also comes with a true label $y_s \in \Y$.
We denote the unlabeled nodes by $\unlbld = V \setminus \seed$,
and set  $n=|V|$, $m=|E|$, $L=|\Y|$, and $k=|\seed|$.
In some settings additional
node features are available. 
We focus on the transductive setting,
where the goal is to predict the labels of
all non-seed nodes $u \in \unlbld$.

The core idea of our method is to propagate labels
from labeled to unlabeled nodes using infection dynamics.
The process is initialized with all seed nodes in an infected state
and all unlabeled nodes in a null state $\nullstate$.
Then, a stochastic model of infection dynamics is used to
determine how the infectious state of nodes in the graph changes over time,
typically as a function of the states of neighboring nodes.
To support multiple label classes,
we consider \emph{competitive} infection models.
In these, seeds $s \in \seed$ are initially infected with their true labels $y_s$,
and compete in infecting unlabeled nodes.

The models we consider are stochastic and converge to a steady state.
This means that, after some time point, the labels
of all nodes will not change anymore (we refer to this as process termination, or steady state).
Since the process itself is stochastic, each instantiation will result in a different value for the labels at termination.
For a given infection model, 
let $\rv_{v\ell}$ be the binary random variable indicating 
whether node $v$ is infected by label $\ell$ at steady state.\footnote{
	For multilabel tasks, $\Y$ is the set of seed node identities, and
	$\f$ becomes a weighted sum of their labels.}
Since our goal is to reason about the labels of the nodes,
it will be natural to utilize the infection dynamics to
generate probabilistic predictions.
For each node $v$, our method outputs
a distribution over labels $\f_v$.
Each entry $\f_{v\ell}$ corresponds to the probability
that $v$ had value $\ell$ at steady state, as a function of the seed set $\seed$ and its labels:
\begin{equation}
\f_{v\ell}(\seed,y) =  \probarg{\rv_{v\ell} = 1}
= \expect{}{\rv_{v\ell}}
\label{eq:prediction_apx}
\end{equation}
Note that $\ell$ can take values in $1,\ldots,L$
but also $\ell=0$ for $\nullstate$.
The entry $\f_{v0}$ therefore describes
the (possibly non-negative) probability that $v$ remained uninfected.

Computing $\f$ exactly is known to be \#P-hard even for simple infection models \cite{chen2010scalable}.
Hence, like many other infection-based methods \cite{kempe2003maximizing,du2013scalable,cohen2014sketch}, 
we resort to a Monte-Carlo approach
and estimate $\f$ by averaging over
infection outcomes $\rv$.
Our final predictor $\fhat$ is:
\begin{equation}
\fhat_{v\ell}(\seed,y) =  
\frac{1}{N} \sum\nolimits_{i=1}^N \rv^{(i)}_{v\ell} 
\label{eq:prediction}
\end{equation}
where $\rv^{(i)}_{v\ell}$ is an indicator for the $i^\text{th}$ random instance.

In principle, outcomes $\rv$ 
can be evaluated by simulating the infection dynamics.
This however is not straightforward for several of the models
we consider, such as those with continuous time.
In the next section we describe some infection models,
and show how $\fhat$ can be efficiently computed for them
using an alternative graphical representation of the infection process.

We conclude by stating an approximation bound for $f$.
As we 
can calculate $\hat{f}$ efficiently (see next section) this implies that our method yields an efficient approximation 
scheme for the true infection probabilities. 

\begin{prop}
For every $\epsilon, \, \delta \in [0,1]$,
if $N\ge \frac{1}{2\epsilon^2} \log{\frac{2 n (L+1)}{\delta}}$,
then with probability of at least $1-\delta$,
Algorithm \ref{algo:infprop} returns $\fhat$
such that $\|\fhat - \f\|_{\max{}} \le \epsilon$.
\thlabel{thm:convergence}
\end{prop}

\begin{proof}
Note that each $\rv_{v\ell}$ is a random variable in
$\{0,1\}$. Furthermore, $\fhat$ is an average of $\rv$,
and $\f$ is the corresponding expectation.
The result is obtained by
applying the Hoeffding and union bounds.
\end{proof}



\subsection{Competitive Infection Models for Graph Labeling \label{sec:diff_models}}
As mentioned above, our SSL method relies on an infection process where
nodes of the graph are ``infected'' with labels.
There are many variants of infection processes (see \cite{chen2013information});
we describe some relevant ones below. 

\subsubsection{The Independent Cascade model}
Since its introduction in \cite{goldenberg2001talk},
the simple but powerful
Independent Cascade (IC) model
has been used extensively.
The original IC model, briefly reviewed below,
is a discrete-time, network-dependent interpretation 
of the classic Susceptible-Infected-Recovered (SIR)
epidemiological model \cite{sir}.
At time $t=0$, seed nodes are initialized
to an infected state,
and all other nodes to a susceptible state.
If node $u$ is infected at time step $t$, then at time $t+1$
it attempts to infect each of its non-infected out-neighbors $v \in \nei{u}$,
and succeeds with
probability $p_{uv}$.
If successful, we refer to the edge $(u,v)$ as \emph{active}
or \emph{activated},
mark the infection time of $v$ as $\tau_v = t+1$,
and set $v$'s \emph{infector} to be $u$, which we denote by $\parent(v) = u$.
The model is therefore parametrized by the set of all edge infection probabilities
$\{ p_{uv} \, | \, (u,v) \in E \}$ (given as input via  $p_{uv} = W_{uv}$).
Once a node becomes infected,
it remains in this state.
As infections are probabilistic,
not all nodes are necessarily infected.
The process terminates
either when all nodes are infected,
or (more commonly) when all infection
attempts at some time step are unsuccessful.


The IC model describes the propagation of a \emph{single} infectious content.
Hence, it can tell us only \emph{when} and \emph{how}
a node is infected, but not by \emph{what}.
This motivated a class of \emph{competitive} infection models
which support multiple content types.
Several competitive IC variants have been proposed \cite{borodin2010threshold,budak2011limiting,chen2011influence,he2012influence}.
The common theme in these is that nodes inherit the label
of their earliest infector (with tie-breaking when needed).
All of these are supported by our method.
In the supplementary material we show how
our approach can also be applied to threshold models \cite{kempe2003maximizing}.

\subsubsection{Continuous Time Dynamics} \label{sec:ctic}

While simple and elegant, the IC dynamics are somewhat limited
in their expressive power.
One important generalization is the Continuous-Time IC model
(CTIC) \cite{mgr2010inferring}.
This model is well suited for SSL as it is flexible,
does not require tie-breaking, and allows for incorporating node priors.
In this model, a successful infection attempt 
entails an ``incubation period'', after which 
the node becomes infected.
Hence, if $u$ succeeds in infecting $v$ at time $\tau_u \in \R^+$,
it draws an incubation time $\delta_{uv} \sim \dist(\theta_{uv})$,
and $v$ can become infected at time $\tau_{u v} = \tau_u + \delta_{uv}$.
As in the IC model, $v$ inherits the label of its earliest infector
$\parent(v) = \argmin_u \tau_{u v}$.
The competitive CTIC model generalizes the competitive IC model
for an appropriate choice of $D$, where
$\delta_e$ is set to 1 with probability $p_e$,
and $\infty$ with probability $1-p_e$.
We therefore consider a general mixture distribution
of activations and incubation times $\dist(p,\theta)$,
where $\delta_e$ is sampled w.p. $p_e$,
and set to $\infty$ w.p. $1-p_e$.
Since infections are determined by the earliest successful attempt,
the shortest-paths interpretation and algorithm (\secref{sec:short_paths})
hold for the random graph $G^\delta = (V,E,\delta)$.


\subsection{Computing Infections Efficiently} \label{sec:algorithm}
For infection models as in \secref{sec:diff_models}, we would like to calculate
predictions $\hat{f}$ 
as in \eqref{eq:prediction}. A naive approach would be to do this by simulating the infection process $N$ times
and averaging.
This, however, is inefficient for discrete-time IC,
requires continuous time simulation for CTIC, and does not apply to general models.
We hence provide an equivalent efficient alternative below.

\begin{figure}[t!] 
\begin{algorithm}[H]
\caption{\textsc{BasicInfProp}($G,\seed,y,p,N$)}
\begin{algorithmic}[1]
  \FOR {$i = 1,\dots,N$}
    \STATE {Initialize $Y^{(i)}_{u\ell} \gets 0$  for all $u \in \unlbld, \, \ell \in \Y \cup \nullstate$}
	\FOR {$(u,v) \in E$}
	  \STATE {$W_{uv} \gets 1$ with probability $p_{uv}$, and $\infty$ o.w.}
	\ENDFOR
	\FOR {$s \in \seed$}
	  \STATE {$\var{dist}[s][\,\cdotp] \gets \method{Dijkstra}(G,W,s)$}
	\ENDFOR
	\FOR {$u \in \unlbld$}
	  \STATE {$Y^{(i)}_{u,\ancestor(u)} \gets 1$ where $\ancestor(u) \in \argmin_{s} \var{dist}[s][u]$}
	\ENDFOR
  \ENDFOR
	\STATE Return $\fhat = \frac{1}{N} \sum_{i=1}^N Y^{(i)}$
\end{algorithmic}
\label{algo:basic_infprop}
\end{algorithm}
\end{figure}  


\subsubsection{Infections as Shortest Paths} \label{sec:short_paths}
We now present an alternative view of the sampling process, which facilitates efficient implementation and extensions.
Consider first the discrete time IC process. For a single instantiation of the process,
recall that if $u$ succeeded in infecting $v$,
the edge $(u,v)$ is considered \emph{active}.
We use the set of active edges $A \subseteq E$ (sampled throughout the instantiation until termination)
to construct the \emph{active graph}
$G^A = (V,E,W^A)$ with
weights $W^A_e = 1$ for $e \in A$
and $W^A_e = \infty$ for $e \in E \setminus A$.
An important observation is that node $v$ is infected at termination iff there exists
a path in $G^A$ from \emph{some} seed node $s \in \seed$ to $v$
with finite weight.
We refer to this as an \emph{active path}.
Since $v$'s actual infection time $\tau_v$
is set by the earliest successful infection,
it is also
the length of the shortest active path from some $s \in \seed$.


The above formulation allows for replacing time with graph distances.
Let $d_{A}(u,v)$ be the distance from $u$ to $v$ in $G^A$.
Due to the recursive nature of label assignment,
it follows that $v$ inherits its label from the $s \in \seed$
whose distance to $v$ is shortest.
We refer to $s$ as $v$'s \emph{ancestor}, denoted by $\ancestor(v)$,
and set $\ancestor(v)=\nullstate$ when there are no paths from $\seed$ to $v$.
Infection outcomes $\rv_{v\ell}$ can now be expressed using distances: 
\begin{equation}
\rv_{v\ell}= \one\{\ell = y_{\ancestor(v)} \}, \qquad
\ancestor(v) = \argmin_{s \in \seed} d_{A}(s,v)
\label{eq:y_as_sp}
\end{equation}

Recall that our motivation here was to compute $Y$ without simulating the dynamics.
Since distances $d_A$ depend on edge activations, it is not yet clear
why \eqref{eq:y_as_sp} is useful.
An important result by \cite{kempe2003maximizing} shows that
ancestors can be computed over a simpler random graph model.
Specifically, let $\Atilde \subseteq E$ be a random edge set,
where \emph{each edge} $(u,v) \in E$ is sampled \emph{independently}
to be in $\Atilde$ with probability $p_{uv}$.
Then, for an appropriately defined $G^{\Atilde}$ and $d_{\Atilde}$,
we have:
\begin{equation}
\ancestor(v) = 
\argmin_{s \in \seed} d_{\Atilde}(s,v)
\label{eq:active_indep}
\end{equation}
Thus, to compute each $\rv_{v\ell}^{(i)}$ (and hence $\fhat$),
it suffices to sample edges independently,
and compute shortest paths on $G^{\Atilde}$,
bypassing the need for simulation.
Under this view, $\f$ can be thought of as an ensemble of shortest-path predictors,
whose weights are set by the dynamics.
Algorithm \ref{algo:basic_infprop} provides a simple implementation of this idea
for the discrete time IC model.
After sampling edges,
the algorithm computes shortest paths (using Dikjstra)
from each $s \in \seed$ to all $u \in \unlbld$.
Then, each node $u$ is assigned the label of its ancestor $\ancestor(u)$.
This approach applies to a large class of infection models
that admit to a similar graphical form \cite{kempe2003maximizing}.





\subsubsection{Improved Efficiency via Modified Dijkstra}
Recall that for a single infection instance,
a node inherits its label from the closest seed node.
Based on this, Algorithm \ref{algo:basic_infprop} offers a direct approach for computing $\f$,
where shortest paths are computed from each of the $k$ seed nodes to every
unlabeled node $v \in \unlbld$
using $k$ calls to Dijkstra.
While correct, this method suffers an unnecessary factor
of $k$ on its runtime.
To reduce this overhead, we change Dijkstra's
initialization and updates, so that only a single call would suffice.
Algorithm \ref{algo:infprop} 
implements this idea for the general CTIC model (\secref{sec:ctic})
and allows for node priors (\secref{sec:extensions}).
The correctness of the algorithm is stated below, and a proof is provided in the  supplementary material.
\begin{prop}
Algorithm \ref{algo:infprop} correctly computes the
estimated infection probabilities $\fhat$ in \eqref{eq:prediction_apx}.
\thlabel{thm:correctness}
\end{prop}
The worst-case complexity of Dijkstra,
and hence of each iteration in Algorithm \ref{algo:infprop}, is
$O(m+n\log n)$. 
Other implementations of Dijkstra which support
further parallelization or GPUs \cite{martin2009cuda}
can also be modified for our setting.
Nonetheless, the practical run time of Algorithm \ref{algo:infprop}
can be, and typically is, much better,
for two reasons.
First, note that only the subset of active edges are traversed
(and sampled on the fly),
and only nodes which are reachable from 
$\seed$ are processed.
The infection parameters $p$ therefore induce a tradeoff 
between the influence diameter of $\seed$
and the run time (empirical demonstration in Fig. \ref{fig:conv_to} (left)).
Second, many settings require ``hard'' predictions $\yhat \in \Y$,
typically set by $\yhat_v = \argmax_\ell \fhat_{u\ell}$.
Hence, for $\yhat_v$ to be correct, it suffices that
$\fhat_{u,y_u} \ge \fhat_{u\ell}$ for all $\ell \in \Y$, which 
does not require the full convergence stated in  \thref{thm:convergence}
(empirical demonstration in Fig. \ref{fig:conv_to} (right)).

In this section we showed how infection outcomes can be computed efficiently.
It is therefore only natural to ask - what is it that infections optimize?
In the next section we show that $\f$
is in fact the solution to a quadratic optimization objective,
whose weights intricately depend on the infection dynamics.




\section{What do infections optimize?} \label{sec:theory}
Many SSL methods propose an optimization objective which
encodes some notion of smoothness.
For instance, the classic LabelProp algorithm \cite{labelprop}
encourages adjacent nodes to agree on their predicted labels
by minimizing a quadratic penalty term:
\begin{equation}
f_\lp = \argmin_{f'} \sum_\ell \sum_{u,v} W_{uv} ( f'_{u\ell} - f'_{v\ell} )^2
\label{eq:labelprop_obj}
\end{equation}
for predictions $\f'$ and symmetric weights $W$,
subject to $\f'_{s\ell} = \1{\ell=y_s}$ for all $s \in \seed$.
In this section we show that InfProp
has a related interpretation.
Specifically, we show that the InfProp predictions $\f$ minimize the quadratic
objective in \eqref{eq:infprop_obj_bias}. 

\comment{
up to bias terms, we show that $\f_\ip$
minimizes the objective:
\begin{equation}
\f_\ip = \argmin_{f'} \sum_\ell \sum_u
\Big( f'_{u\ell} - \sum_v w_{uv}^{(\seed)} f'_{v\ell} \Big)^2
\label{eq:infprop_obj}
\end{equation}
}


\begin{figure}[t!] 
\begin{algorithm}[H]
	\caption{\textsc{InfProp}
		($G,\seed,y,\dist,\pnlty,N$)} 
	\begin{algorithmic}[1]
    \FOR {$i = 1,\dots,N$}
      \STATE {Initialize $Y^{(i)}_{u\ell} \gets 0$  for all $u \in \unlbld, \, \ell \in \Y \cup \nullstate$}
      \FOR {$v \in \unlbld$}
      \STATE $\var{dist}[v] \gets \infty, \,\,\, \var{y}[v] \gets \nullstate$ 
      \ENDFOR
      \FOR {$s \in \seed$}
      \STATE $\var{dist}[s] \gets 0, \,\,\, \var{y}[s] \gets y_s$
      \STATE push $s$ into min-queue $Q$
      \ENDFOR
      \WHILE{$Q$ is not empty}
      \STATE pop $v$ from $Q$ \CMNT{break ties randomly}
      \FOR {$u \in \nei{v}$}
      \STATE sample $\delta_{vu} \sim D(\theta,p)$ \CMNT{incubation time}
      \STATE \textbf{if} $\delta_{vu}=\infty$ \textbf{then} continue 
      \STATE $\var{alt} \gets \var{dist}[v] + w^A_{vu} + \pnlty_{u}(\var{y}[v])$  \CMNT{penalize}
      \label{algo_line:line:update}
      \IF{$\var{alt} < \var{dist}[u]$}
      \STATE $\var{dist}[u] = \var{alt}$
      \STATE $\var{y}[u] \gets \var{y}[v]$ \CMNT{$u$ inherits label from parent $v$}
      \STATE push/update $u$ in $Q$ with $\var{dist}[u]$
      \ENDIF
      \ENDFOR
      \ENDWHILE
      \STATE $Y^{(i)}_{u,\var{y}[v]} \gets 1$ for all $u \in \unlbld$
    \ENDFOR
    \STATE Return $\fhat = \frac{1}{N} \sum_{i=1}^N Y^{(i)}$
	\end{algorithmic}
	\label{algo:infprop}
\end{algorithm}
\end{figure}


While similar in structure,
the fundamental difference between
Eqs. (\ref{eq:labelprop_obj}) and (\ref{eq:infprop_obj_bias})
lies in how the weights 
are determined.
In LabelProp (and variants),
edge weights are given as input, and are typically
set according to some feature-based similarity measure.
In this sense, each $W_{uv}$ is a local function
of the features of $u$ and $v$.
In contrast,
weights in \eqref{eq:infprop_obj_bias} are set in a global manner.
As we show next, 
each  weight is a function of the infection dynamics,
of the specific seed set $\seed$,
and, if available, of the features of \emph{all} nodes.
To demonstrate this, and to see why \eqref{eq:infprop_obj_bias} holds,
it will be helpful to analyze InfProp from a spectral perspective.

\subsection{A Laplacian Interpretation for InfProp}
An interesting property of LabelProp
is that its objective can be expressed via the graph Laplacian.
For a directed weighted graph, 
the normalized Laplacian is: 
\begin{equation}
\lap_\lp = I - \widetilde{W}
\label{eq:laplacian}
\end{equation}
where $\widetilde{W} = D^{-1} W$, 
$D$ is diagonal with $D_{uu} = \sum_v W_{uv}$
(and $W$ is symmetric).
The output of LabelProp can be computed by solving the system
$\lap_\lp \f' = 0$ for the unlabeled nodes.
We now show that the infection-based predictions of InfProp
also correspond to the solution of a certain Laplacian system
which is determined by the seed set and the infection dynamics.

Consider a single infection instance,
and denote by $\parmat_{uv}(\seed)$ the random variable indicating
whether $u$ was infected by $v$ for seed $\seed$,
namely $\parmat_{uv}(\seed) = \1{u = \parent(v)}$.
We refer to the matrix $\parmat$ as the \emph{infector matrix}.
Further denote by $\expparmat$ the expected infector matrix 
$\expparmat(\seed) = \expect{}{\parmat(\seed)}$.
We use this to define the following Laplacian:
	\begin{equation}
\lap(\seed)_\ip = I-\expparmat(\seed)
\label{eq:infprop_laplacian}
\end{equation}
Note that $\lap_\ip$ is defined over the same graph $G$, 
but need not be symmetric.
We now show that $\lap$ is indeed a Laplacian matrix, 
and that it can be used to infer $\f$.


\begin{lemma} \thlabel{thm:lap_nonhom}
	The infection-based predictions $\f_\ip$ in \eqref{eq:prediction}
	are also the solution to the
	Laplacian system:
	\begin{equation}
	\lap(\seed)_\ip \f = \Bias(\seed) 
	\label{eq:laplacian_bias_eq}
	\end{equation}
	where:
	\[
	\Bias_{u\ell}(\seed) = \sum_v b_{vu\ell}^{(\seed)},
	\qquad
	b_{vu\ell}^{(\seed)} = \covargs{}{\parmat_{vu}(\seed),Y_{u\ell}}
	\]
\end{lemma}
For conciseness, we defer the full proof to the supplementary material,
and show here a useful special case.

\begin{lemma}	\thlabel{thm:lap_hom}
	If $\parmat$ and $Y$ are uncorrelated,
	then the infection-based predictions $\f_\ip$ in \eqref{eq:prediction}
	are also the solution to the homogeneous Laplacian system:
	\begin{equation}
	\lap_{\ip}(\seed) f = 0
		\label{eq:laplacian_hom_eq}
	\end{equation}
\end{lemma}

\begin{proof}
We first show that $\lap$ is a graph Laplacian,
namely that the sum of each row in $\expparmat$
is equal to the corresponding diagonal element in $I$, which is 1.
Since rows in $\parmat$ have only one non-zero entry of value one,
each row in $\expparmat$ is positive and sums to one.
Note that $\expparmat_{u\cdotp}$ provides a distribution over the infectors of $u$.

We now prove \eqref{eq:laplacian_hom_eq}.
By definition, the label of each node at steady state
is set to be that of its infector,
namely
$Y_{v\ell} = Y_{\parent(v),\ell}$ for all $v$ and $\ell$,
or simply $Y=TY$.
Using \eqref{eq:prediction} and applying expectation, we have:
\begin{equation}
\f(\seed)_\ip = \expect{}{Y} = \expect{}{\parmat(\seed) Y}
\label{eq:recursive_expect}
\end{equation}
When $\parmat,Y$ are uncorrelated,
$\expect{}{\parmat Y} = \expect{}{\parmat} \expect{}{Y}$,
hence $\f_\ip = \expparmat \f_\ip$.
Rearranging 
concludes our proof.
\end{proof}

\subsection{InfProp as Optimization}
We next use the Laplacian insight above to provide an objective minimized by the InfProp
solution. 
Begin by noting that for LabelProp,
the solution of \eqref{eq:laplacian} coincides with
the solution of the following objective:
\begin{align} 
\f_\lp &= \argmin_{f'} \| \lap_\lp f'  \|_F^2 \nonumber \\
&= \argmin_{f'} \sum_\ell \sum_u \Big( f'_{u\ell} - \sum_v \widetilde{W}_{uv} \f'_{v\ell} \Big)^2  \label{eq:lp_obj_norm}
\end{align}
where minimization is only over the unlabeled nodes,
and $\|\cdotp\|_F$ denotes the Frobenius norm.
This gives an alternative quadratic objective
which bounds \eqref{eq:labelprop_obj} 
and directly expresses the steady-state of LabelProp's averaging dynamics.
In a similar fashion,
we can derive an equivalent formulation of $f$ in \eqref{eq:laplacian_bias_eq} via:
\begin{equation}
\f(\seed) = \argmin_{f'} \| \lap(\seed)_\ip f' - \Bias(\seed) \|_F^2
\label{eq:infprop_obj_norm}
\end{equation}
Expanding and denoting $w_{uv}^{(\seed)} = \expparmat_{uv}(\seed)$
provides the general objective of our method:
\begin{equation}
\min_{f'} \sum_\ell \sum_u
\Big( f'_{u\ell} - \sum_v
\big( w_{uv}^{(\seed)} f'_{v\ell} + \bias_{uv\ell}^{(\seed)} \big) \Big)^2
\label{eq:infprop_obj_bias}
\end{equation}

Note that \eqref{eq:infprop_obj_bias} and \eqref{eq:lp_obj_norm}
are structurally equivalent up to the bias terms,
which disappear under the conditions of \thref{thm:lap_hom}.
The critical difference is that the weights in \eqref{eq:infprop_obj_bias}
are now \emph{functions} of the dynamics and seed set,
rather than just scalars given as input. 
Through their dependence on $\expparmat$ and $Y$,
the weights and bias terms in \eqref{eq:infprop_obj_bias} are in fact functions of the dynamics.
In this sense, $w_{uv}^{(\seed)}$ quantifies how well $v$ relays
information from $\seed$ to $u$, which depends on the entire graph.
Similarly,
the term $\bias_{uv\ell}^{(\seed)}$ quantifies consistency
between the identity of $u$'s infector ($v$) and the inherited label ($\ell$).
This means that frequent yet indecisive infectors are penalized,
while reliable nodes remain unbiased. 

Finally, note that the optimization interpretation above does not offer a better optimization scheme, since calculating the weights $w(\seed)$ and $\bias(\seed)$ would require sampling.
Hence, our InfProp sampling algorithms from \secref{sec:algorithm} would be a simpler approach.

\comment{
Note that unlike LabelProp,
an optimizational perspective does not provide any computational advantages for our method.
This is for several reasons.
First, optimizing an inhomogeneous directed Laplacian system such as in
\eqref{eq:laplacian_bias_eq}
is a notoriously difficult task \cite{vishnoi2013Lxb}.
Second, just as any quadratic objective,
the a-symmetric weights in \eqref{eq:infprop_obj_bias}
can cause the objective to be non-convex.
Third, since the weights depend on infection outcomes,
computing them can be as hard as computing $\f$ itself.
Nonetheless, \secref{sec:algorithm} shows that $\f$ can still be approximated
to within arbitrary precision.
}


\begin{figure*}[t]
\begin{center}
\includegraphics[width=\textwidth]{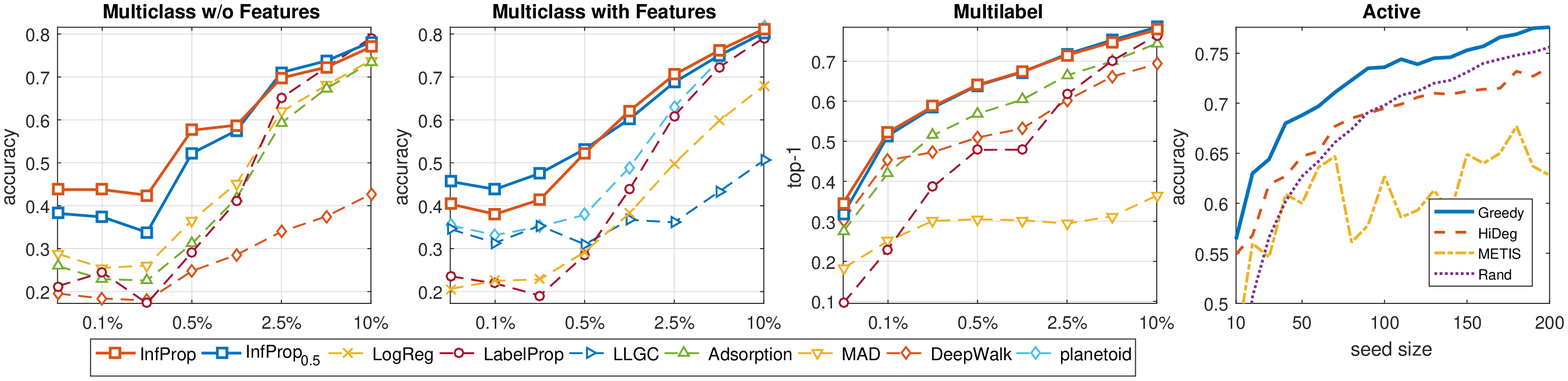}
\caption{Results on the CoRA dataset for various learning settings.}
\label{fig:cora}
\end{center}
\end{figure*}


\section{Other Learning Settings} \label{sec:extensions}
In this section we briefly describe how our method
extends to other learning settings used in our experiments.
For more details please see the supp. material.

\paragraph{Incorporating features and priors:}
Many network-based datasets include additional node features or priors.
Our method incorporates priors directly into the CTIC dynamics
by penalizing incubation times.
Denote by $\prior_{v\ell}$ the prior for labeling $v$ with $\ell$,
and let $\pnlty : [0,1] \rightarrow \R$ be a penalty function.
If $u$ succeeds in infecting $v$ with $\ell$,
the incubation time $\delta_{uv}$ is penalized by an additional
$\pnlty(\prior_{v\ell})$.
For a monotone decreasing $\pnlty$, high priors induce low penalties, and vice versa.
Although penalties are deployed locally,
they delay the propagation of the penalized label across the graph in a global manner.


\paragraph{Confidence and active learning:}
Recall that $v$ remains uninfected with probability $f_{v0}$.
Hence, $\conf_v(\seed) = 1 - \f_{v0}$ serves as a natural measure of confidence.
We use this as a selection criteria for an active setting
where the goal is to choose a seed set of size $k$.
The objective we consider coincides with the well-studied notion of 
influence \cite{kempe2003maximizing},
which is monotone and submodular and admits to an efficient greedy approximation scheme.
Our method thus offers a tractable alternative
to existing active SSL methods 
\cite{guillory2012active,gadde2014active,gu2012towards}.


\section{Related Work \label{sec:related}}
Methods for SSL are often based on assumptions regarding the
structure of the unlabeled data.
One such assumption is \emph{smoothness},
which states that examples that are close are likely to have similar labels.
In the classic Label Propagation algorithm \cite{labelprop},
adjacent nodes in the graph are encouraged to agree on their labels
via a quadratic penalty.
Some variants
add regularization terms \cite{adsorption},
allow for label uncertainty \cite{mad},
or include normalization and unanchored seeds \cite{llgc}.

The above methods are designed for graphs that approximate
the data density via similarity in feature space, and are typically
constructed from samples.
Recent SSL methods are geared towards tasks where graphs are an
additional part of the input.
Motivated by deep embeddings \cite{mikolov2013distributed},
these methods embed the nodes of a graph into a low-dimensional vector space, which can then be used in various ways.
When the data includes only the graph,
the embeddings can be used as input
for an off-the-shelf predictor \cite{deepwalk}.
When the data includes additional node features,
the embedding can act as a regularizer for a
standard loss over the labeled nodes
\cite{planetoid,gcn}.
In contrast to classic methods,
these methods propagate \emph{features} rather than labels.


\begin{figure}[!b]
\begin{center}
\includegraphics[width=\columnwidth]{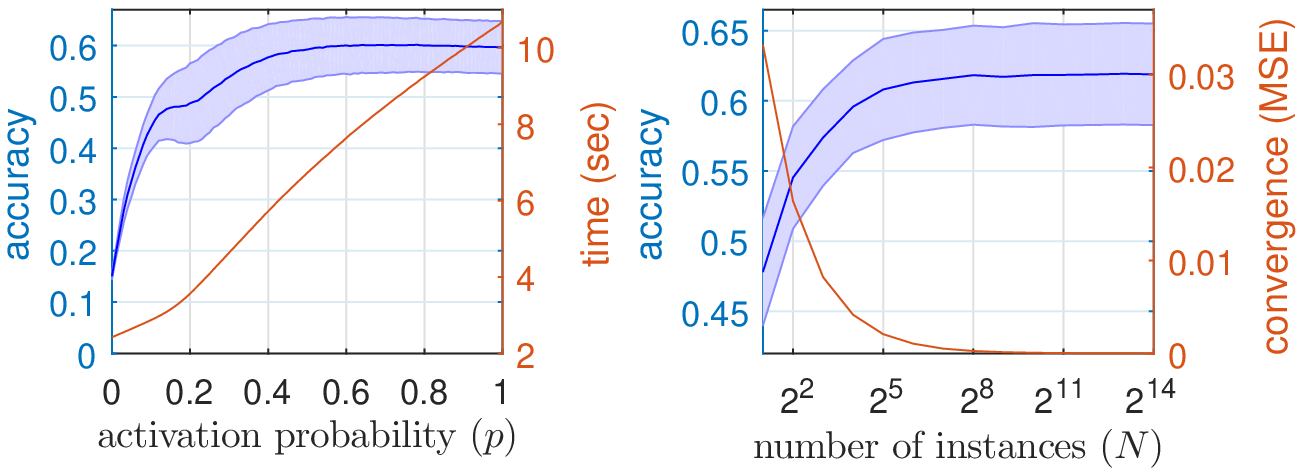}
\captionof{figure}{Activation tradeoff and convergence}
\label{fig:conv_to}
\end{center}
\end{figure}

An alternative method for utilizing graphs is to consider
shortest paths as a measure of closeness.
The authors of \cite{alamgir2011phase} show that Laplacians and shortest
paths are special cases of ``resistance distances'',
and propose (but do not evaluate) a new regularizer.
Other methods construct ad-hoc graphs whose shortest paths
approximate density-based distances \cite{orlitsky2005estimating,bijral2011semi}.
A recent work \cite{cohen2016semi} proposes
a method for SSL in directed graphs based on distance diffusion.
As they consider distances from unlabeled to labeled nodes,
each instance is computationally intensive,
and requires an approximation scheme.
In contrast, we consider distances from labeled to unlabeled nodes,
which can be computed efficiently.
While for a specific setting (symmetric weights and a certain link function)
both models overlap,
in this paper we consider a more general setup.

Our method draws on the rich literature of
infection models and diffusion processes over networks.
These have been used for describing the propagation of information,
innovation, behavioral norms, and others,
and have been utilized in works in influence maximization \cite{kempe2003maximizing},
network inference \cite{mgr2010inferring},
influence maximization \cite{kempe2003maximizing}
estimation \cite{du2013scalable,cohen2014sketch}
and prediction \cite{dimple},
and personalized marketing \cite{chen2010scalable}.



\begin{table*}[t]
\setlength{\tabcolsep}{3pt}
  \centering
  \tblsmall
          \begin{tabular}{|l|ccccc|cccccc|}
    \multicolumn{1}{r}{} & \multicolumn{5}{c}{Multiclass (Accuracy\,/\,MSE)}       & \multicolumn{5}{c}{Multilabel (AUC\,/\,Top-1)}       & \multicolumn{1}{r}{} \bigstrut[b]\\
\cline{2-12}    \multicolumn{1}{r|}{} & CoRA  & DBLP  & Flickr & IMDb  & Industry & Amazon & CoRA  & IMDb  & PubMed & Wikipedia & YouTube \bigstrut\\
    \hline
InfProp & \textbf{0.59}\tbltiny{\,/\,\textbf{0.56}} & 0.73\tbltiny{\,/\,\textbf{0.42}} & \textbf{0.79}\tbltiny{\,/\,\textbf{0.38}} & 0.56\tbltiny{\,/\,\textbf{0.49}} & \textbf{0.21}\tbltiny{\,/\,0.91} & 0.79\tbltiny{\,/\,0.56} & \textbf{0.91}\tbltiny{\,/\,\textbf{0.67}} & 0.75\tbltiny{\,/\,\textbf{0.36}} & \textbf{0.90}\tbltiny{\,/\,\textbf{0.77}} & 0.93\tbltiny{\,/\,0.70} & \textbf{0.84}\tbltiny{\,/\,0.38} \bigstrut[t]\\
InfProp$_{0.5}$ & 0.58\tbltiny{\,/\,0.64} & 0.74\tbltiny{\,/\,0.46} & 0.78\tbltiny{\,/\,\textbf{0.38}} & 0.55\tbltiny{\,/\,\textbf{0.49}} & \textbf{0.21}\tbltiny{\,/\,\textbf{0.90}} & 0.79\tbltiny{\,/\,\textbf{0.57}} & 0.90\tbltiny{\,/\,0.67} & 0.76\tbltiny{\,/\,\textbf{0.36}} & 0.88\tbltiny{\,/\,\textbf{0.77}} & \textbf{0.94}\tbltiny{\,/\,0.71} & \textbf{0.84}\tbltiny{\,/\,\textbf{0.40}} \\
ShortPaths & 0.53\tbltiny{\,/\,0.87} & 0.63\tbltiny{\,/\,0.74} & 0.65\tbltiny{\,/\,0.70} & 0.55\tbltiny{\,/\,0.90} & 0.16\tbltiny{\,/\,1.57} & 0.69\tbltiny{\,/\,0.49} & 0.76\tbltiny{\,/\,0.56} & 0.57\tbltiny{\,/\,0.30} & 0.76\tbltiny{\,/\,0.67} & 0.67\tbltiny{\,/\,0.36} & 0.59\tbltiny{\,/\,0.22} \\
LabelProp & 0.41\tbltiny{\,/\,0.74} & 0.60\tbltiny{\,/\,0.59} & 0.33\tbltiny{\,/\,0.90} & 0.50\tbltiny{\,/\,0.63} & 0.14\tbltiny{\,/\,0.99} & \textbf{0.85}\tbltiny{\,/\,\textbf{0.57}} & 0.86\tbltiny{\,/\,0.48} & \textbf{0.77}\tbltiny{\,/\,\textbf{0.36}} & 0.82\tbltiny{\,/\,0.64} & 0.78\tbltiny{\,/\,0.31} & 0.71\tbltiny{\,/\,0.18} \\
Adsorption & 0.42\tbltiny{\,/\,0.99} & 0.54\tbltiny{\,/\,0.99} & 0.72\tbltiny{\,/\,0.99} & 0.56\tbltiny{\,/\,0.99} & 0.14\tbltiny{\,/\,0.99} & 0.73\tbltiny{\,/\,0.52} & 0.86\tbltiny{\,/\,0.60} & 0.71\tbltiny{\,/\,0.32} & 0.80\tbltiny{\,/\,0.69} & 0.89\tbltiny{\,/\,0.57} & 0.79\tbltiny{\,/\,0.28} \\
MAD   & 0.45\tbltiny{\,/\,0.99} & 0.20\tbltiny{\,/\,1.00} & 0.75\tbltiny{\,/\,0.99} & \textbf{0.58}\tbltiny{\,/\,0.99} & 0.16\tbltiny{\,/\,1.00} & 0.73\tbltiny{\,/\,0.52} & 0.47\tbltiny{\,/\,0.30} & 0.70\tbltiny{\,/\,0.32} & 0.79\tbltiny{\,/\,0.70} & 0.00\tbltiny{\,/\,0.05} & 0.81\tbltiny{\,/\,0.31} \\
DeepWalk & 0.29\tbltiny{\,/\,0.86} & \textbf{0.77}\tbltiny{\,/\,0.62} & 0.49\tbltiny{\,/\,0.73} & 0.50\tbltiny{\,/\,0.56} & 0.17\tbltiny{\,/\,0.92} & 0.60\tbltiny{\,/\,0.13} & 0.80\tbltiny{\,/\,0.53} & 0.61\tbltiny{\,/\,0.32} & 0.57\tbltiny{\,/\,0.42} & \textbf{0.94}\tbltiny{\,/\,\textbf{0.88}} & 0.60\tbltiny{\,/\,0.24} \bigstrut[b]\\
    \hline
    \end{tabular}%
		\caption{Results for experiments on data without features.}
  \label{tbl:mc_ml}%
\end{table*}%

\section{Experiments} \label{sec:experiments}
We evaluated our method on various learning tasks
over three benchmark dataset collections, which include networked data for
multiclass learning with features 
\cite{linqs}
and  without features 
\cite{flip}, 
and multilabel learning 
\cite{snbc}. 
The datasets include diverse networks such as
social networks, citation and co-authorship graphs,
product and item networks, and hyperlink graphs
(see supplementary material for dataset summary statistics).

Our experimental setup follows the standard 
graph-based semi-supervised learning evaluation approach.
Specifically, in each instance we draw a seed set
of size $k$ uniformly at random, acquire its labels,
and then use the graph and labeled seed set
to generate labels for all nodes.
We repeat this procedure for 10 random seed set selections
and for various values of $k$
(where $k$ is set
to be a fixed proportion of the number of nodes in the graph)
and report average results.

We compared our method to current state-of-the-art
baselines, which include spectral methods as well as
deep embedding methods.
For tasks which do not include features, these included
\method{LabelProp} \cite{labelprop},
\method{Adsorption} \cite{adsorption}, \method{MAD} \cite{mad},
and the feature-agnostic deep method \method{DeepWalk} \cite{deepwalk}.
For tasks which do include features, we compared to 
the prior-supporting spectral method \method{LLGC} \cite{llgc},
the recent feature-based deep method \method{Planetoid} \cite{planetoid},
\method{LabelProp} as a graph-only baseline,
logistic regression (\method{LogReg}) as a features-only baseline,
and a baseline where labels are set by shortest paths in $G$ (\method{ShortPaths}).
For the active setting (Fig. \ref{fig:cora}), we compared our 
approach (\method{Greedy}) to \method{METIS} \cite{guillory2009label},
to choosing high-degree nodes (\method{HiDeg}),
and to random seeds (\method{Rand}).

For our method (\method{InfProp})
we used exponential
incubation times $\delta \sim \text{Exp}(\theta)$.
As in many works (e.g., \cite{kempe2003maximizing,cohen2016semi}),
we used
$\theta_{uv} = 1/d_u$ 
for all node pairs $(u,v) \in E$,
where $d_u$ is the out-degree of $u$.
We set the number of random instances to $N=1,000$.
Fig. \ref{fig:conv_to} (right) demonstrates accuracy and convergence
as a function of $N$.
We show results for two variants:
\method{InfProp}, where we set activation probabilities to $p=1$ for all edges,
and \method{InfProp}$_{0.5}$, where $p=0.5$.
In addition to providing a confidence measure,
\method{InfProp}$_{0.5}$ is much faster,
while on average achieving 0.99\% of the performance of \method{InfProp}. 
Fig. \ref{fig:conv_to} (left) demonstrates the tradeoff in accuracy
and runtime when varying $p$.

The methods we consider naturally output probabilistic
``soft'' labels as predictions.
We therefore evaluate performance using both probabilistic (for multi-class)
or order-based (for multi-label) performance measures,
as well as performance measures for ``hard'' labels,
which were generated
by choosing the label with the highest value.
Tables \ref{tbl:mc_ml} and \ref{tbl:mcx}
include results for all datasets for $k=1\%$
of the data.
Fig. \ref{fig:cora} shows results
for various values of $k$ on the CoRA dataset (which appears in all benchmarks).
As shown, \method{InfProp} consistently performs well across all settings.

\section{Conclusions}

In this work we presented an SSL method where labels propagate over the graph using dynamic infection models.
These models have a strong connection to short-path ensembles and to
graph Laplacians, allow for efficient computation, and show empirical potential.
Our work was motivated by the idea that
different graph types may require
different dynamics, which
led us to consider alternatives to random walks
and averaging dynamics. 
We used a competitive CTIC variant,
but other infection models (and other dynamics in general)
can be considered.
The choice of dynamics can serve as a means for expressing
prior knowledge and for encoding structure and dependencies.

The models we use have very few tunable parameters.
Nonetheless, one can consider highly parametrized models.
Such parameters can be used to control infection probabilities,
be node or label specific, relate to features,
and even adjust the dynamics themselves.
The stochastic nature of the models and the nonlinearity of the dynamics
makes learning these parameters a challenging task,
which we leave for future work.


\begin{table}[!h]
  \centering
  \fontsize{7.5pt}{9pt}\selectfont
	\tblsmallb
    \begin{tabular}{|l|lll|}
\cline{2-4}    \multicolumn{1}{l|}{(Acc. / MSE)} & CiteSeer & CoRA  & PubMed \bigstrut\\
    \hline
    \method{InfProp} & 0.47\tbltinyb{ / 0.72} & \textbf{0.62}\tbltinyb{ / 0.59} & \textbf{0.74}\tbltinyb{ / 0.46} \bigstrut[t]\\
    \method{InfProp}$_{0.5}$ & \textbf{0.48}\tbltinyb{ / 0.74} & 0.60\tbltinyb{ / \textbf{0.57}} & 0.72\tbltinyb{ / \textbf{0.41}} \\
    \method{ShortPaths} & 0.39\tbltinyb{ / 0.73} & 0.44\tbltinyb{ / 0.72} & 0.68\tbltinyb{ / 0.51} \\
    \method{LogReg} & 0.44\tbltinyb{ / 0.78} & 0.37\tbltinyb{ / 0.81} & 0.45\tbltinyb{ / 0.65} \\
    \method{LabelProp} & 0.39\tbltinyb{ / 0.77} & 0.38\tbltinyb{ / 0.78} & 0.40\tbltinyb{ / 0.67} \\
    \method{LLGC}  & 0.45\tbltinyb{ / \textbf{0.71}} & 0.49\tbltinyb{ / 0.69} & 0.44\tbltinyb{ / 0.67} \\
    \method{Planetoid}\footnote{Results differ from \cite{planetoid} since their evaluation is based on a specific seed, chosen by a different procedure, evaluated on 1000 samples,
    and early-stopped differently.} & 0.41\tbltinyb{ / 0.94} & 0.53\tbltinyb{ / 0.89} & 0.68\tbltinyb{ / 0.64} \bigstrut[b]\\
    \hline
    \end{tabular}%
	\caption{Results on data with features.}
  \label{tbl:mcx}%
\end{table}%



%

\paragraph{Acknowledgments.}
This work was supported by the Blavatnik Computer Science Research Fund
and an ISF Centers of Excellence grant.

\bibliographystyle{acm}
\bibliography{semisupinf}

\begin{thebibliography}{10}

\bibitem{alamgir2011phase}
{\sc Alamgir, M., and Luxburg, U.~V.}
\newblock Phase transition in the family of p-resistances.
\newblock In {\em Advances in Neural Information Processing Systems\/} (2011),
  pp.~379--387.

\bibitem{aleliunas1979random}
{\sc Aleliunas, R., Karp, R.~M., Lipton, R.~J., Lovasz, L., and Rackoff, C.}
\newblock Random walks, universal traversal sequences, and the complexity of
  maze problems.
\newblock In {\em Foundations of Computer Science, 1979., 20th Annual Symposium
  on\/} (1979), IEEE, pp.~218--223.

\bibitem{altun2006maximum}
{\sc Altun, Y., McAllester, D., and Belkin, M.}
\newblock Maximum margin semi-supervised learning for structured variables.
\newblock {\em Advances in neural information processing systems 18\/} (2006),
  33.

\bibitem{adsorption}
{\sc Baluja, S., Seth, R., Sivakumar, D., Jing, Y., Yagnik, J., Kumar, S.,
  Ravichandran, D., and Aly, M.}
\newblock Video suggestion and discovery for youtube: taking random walks
  through the view graph.
\newblock In {\em Proceedings of the 17th international conference on World
  Wide Web\/} (2008), ACM, pp.~895--904.

\bibitem{belkin2004semi}
{\sc Belkin, M., and Niyogi, P.}
\newblock Semi-supervised learning on {R}iemannian manifolds.
\newblock {\em Machine Learning 56}, 1 (2004), 209--239.

\bibitem{bijral2011semi}
{\sc Bijral, A.~S., Ratliff, N., and Srebro, N.}
\newblock Semi-supervised learning with density based distances.
\newblock In {\em Proceedings of the Twenty-Seventh Conference on Uncertainty
  in Artificial Intelligence\/} (2011), UAI'11, pp.~43--50.

\bibitem{blum2001learning}
{\sc Blum, A., and Chawla, S.}
\newblock Learning from labeled and unlabeled data using graph mincuts.
\newblock {\em Proc. 18th International Conf. on Machine Learning\/} (2001).

\bibitem{borodin2010threshold}
{\sc Borodin, A., Filmus, Y., and Oren, J.}
\newblock Threshold models for competitive influence in social networks.
\newblock In {\em WINE\/} (2010), vol.~6484, Springer, pp.~539--550.

\bibitem{broder1994trading}
{\sc Broder, A.~Z., Karlin, A.~R., Raghavan, P., and Upfal, E.}
\newblock Trading space for time in undirected s-t connectivity.
\newblock {\em SIAM Journal on Computing 23}, 2 (1994), 324--334.

\bibitem{budak2011limiting}
{\sc Budak, C., Agrawal, D., and El~Abbadi, A.}
\newblock Limiting the spread of misinformation in social networks.
\newblock In {\em Proceedings of the 20th international conference on World
  wide web\/} (2011), ACM, pp.~665--674.

\bibitem{chapelle2006semi}
{\sc Chapelle, O., Sch{\"o}lkopf, B., Zien, A., et~al.}
\newblock {\em Semi-supervised learning}.
\newblock MIT press Cambridge, 2006.

\bibitem{chen2011influence}
{\sc Chen, W., Collins, A., Cummings, R., Ke, T., Liu, Z., Rincon, D., Sun, X.,
  Wang, Y., Wei, W., and Yuan, Y.}
\newblock Influence maximization in social networks when negative opinions may
  emerge and propagate.
\newblock In {\em Proceedings of the 2011 SIAM International Conference on Data
  Mining\/} (2011), SIAM, pp.~379--390.

\bibitem{chen2013information}
{\sc Chen, W., Lakshmanan, L.~V., and Castillo, C.}
\newblock Information and influence propagation in social networks.
\newblock {\em Synthesis Lectures on Data Management 5}, 4 (2013), 1--177.

\bibitem{chen2010scalable}
{\sc Chen, W., Wang, C., and Wang, Y.}
\newblock Scalable influence maximization for prevalent viral marketing in
  large-scale social networks.
\newblock In {\em Proceedings of the 16th ACM SIGKDD international conference
  on Knowledge discovery and data mining\/} (2010), ACM, pp.~1029--1038.

\bibitem{cohen2016semi}
{\sc Cohen, E.}
\newblock Semi-supervised learning on graphs through reach and distance
  diffusion.
\newblock {\em arXiv preprint arXiv:1603.09064\/} (2016).

\bibitem{cohen2014sketch}
{\sc Cohen, E., Delling, D., Pajor, T., and Werneck, R.~F.}
\newblock Sketch-based influence maximization and computation: Scaling up with
  guarantees.
\newblock In {\em Proceedings of the 23rd ACM International Conference on
  Conference on Information and Knowledge Management\/} (2014), ACM,
  pp.~629--638.

\bibitem{du2013scalable}
{\sc Du, N., Song, L., Rodriguez, M.~G., and Zha, H.}
\newblock Scalable influence estimation in continuous-time diffusion networks.
\newblock In {\em Advances in neural information processing systems\/} (2013),
  pp.~3147--3155.

\bibitem{gadde2014active}
{\sc Gadde, A., Anis, A., and Ortega, A.}
\newblock Active semi-supervised learning using sampling theory for graph
  signals.
\newblock In {\em Proceedings of the 20th ACM SIGKDD international conference
  on Knowledge discovery and data mining\/} (2014), ACM, pp.~492--501.

\bibitem{goldenberg2001talk}
{\sc Goldenberg, J., Libai, B., and Muller, E.}
\newblock Talk of the network: A complex systems look at the underlying process
  of word-of-mouth.
\newblock {\em Marketing letters 12}, 3 (2001), 211--223.

\bibitem{mgr2010inferring}
{\sc Gomez~Rodriguez, M., Leskovec, J., and Krause, A.}
\newblock Inferring networks of diffusion and influence.
\newblock In {\em Proceedings of the 16th ACM SIGKDD International Conference
  on Knowledge Discovery and Data Mining\/} (2010), KDD '10, pp.~1019--1028.

\bibitem{mgr2016influence}
{\sc Gomez-Rodriguez, M., Song, L., Du, N., Zha, H., and Sch\"{o}lkopf, B.}
\newblock Influence estimation and maximization in continuous-time diffusion
  networks.
\newblock {\em ACM Trans. Inf. Syst. 34}, 2 (Feb. 2016), 9:1--9:33.

\bibitem{gu2012towards}
{\sc Gu, Q., and Han, J.}
\newblock Towards active learning on graphs: An error bound minimization
  approach.
\newblock In {\em Data Mining (ICDM), 2012 IEEE 12th International Conference
  on\/} (2012), IEEE, pp.~882--887.

\bibitem{guillory2009label}
{\sc Guillory, A., and Bilmes, J.~A.}
\newblock Label selection on graphs.
\newblock In {\em Advances in Neural Information Processing Systems\/} (2009),
  pp.~691--699.

\bibitem{guillory2012active}
{\sc Guillory, A., and Bilmes, J.~A.}
\newblock Active semi-supervised learning using submodular functions.
\newblock In {\em {UAI} 2011, Proceedings of the Twenty-Seventh Conference on
  Uncertainty in Artificial Intelligence\/} (2011), pp.~274--282.

\bibitem{he2012influence}
{\sc He, X., Song, G., Chen, W., and Jiang, Q.}
\newblock Influence blocking maximization in social networks under the
  competitive linear threshold model.
\newblock In {\em Proceedings of the 2012 SIAM International Conference on Data
  Mining\/} (2012), SIAM, pp.~463--474.

\bibitem{kempe2003maximizing}
{\sc Kempe, D., Kleinberg, J., and Tardos, {\'E}.}
\newblock Maximizing the spread of influence through a social network.
\newblock In {\em Proceedings of the ninth ACM SIGKDD international conference
  on Knowledge discovery and data mining\/} (2003), ACM, pp.~137--146.

\bibitem{sir}
{\sc Kermack, W.~O., and McKendrick, A.~G.}
\newblock A contribution to the mathematical theory of epidemics.
\newblock {\em Proceedings of the Royal Society of London A: Mathematical,
  Physical and Engineering Sciences 115}, 772 (1927), 700--721.

\bibitem{gcn}
{\sc Kipf, T.~N., and Welling, M.}
\newblock Semi-supervised classification with graph convolutional networks.
\newblock {\em arXiv preprint arXiv:1609.02907\/} (2016).

\bibitem{lin2008multirank}
{\sc Lin, F., and Cohen, W.~W.}
\newblock The multirank bootstrap algorithm: Self-supervised political blog
  classification and ranking using semi-supervised link classification.
\newblock In {\em Proceedings of the Second International Conference on Weblogs
  and Social Media, {ICWSM} 2008, Seattle, Washington, USA, March 30 - April 2,
  2008\/} (2008).

\bibitem{lofgren2014fast}
{\sc Lofgren, P., Banerjee, S., Goel, A., and Comandur, S.}
\newblock {FAST-PPR:} scaling personalized pagerank estimation for large
  graphs.
\newblock In {\em The 20th {ACM} {SIGKDD} International Conference on Knowledge
  Discovery and Data Mining, {KDD} '14, New York, NY, {USA} - August 24 - 27,
  2014\/} (2014), pp.~1436--1445.

\bibitem{martin2009cuda}
{\sc Mart{\'\i}n, P.~J., Torres, R., and Gavilanes, A.}
\newblock Cuda solutions for the sssp problem.
\newblock In {\em International Conference on Computational Science\/} (2009),
  Springer, pp.~904--913.

\bibitem{mikolov2013distributed}
{\sc Mikolov, T., Sutskever, I., Chen, K., Corrado, G.~S., and Dean, J.}
\newblock Distributed representations of words and phrases and their
  compositionality.
\newblock In {\em Advances in neural information processing systems\/} (2013),
  pp.~3111--3119.

\bibitem{nadler2009semi}
{\sc Nadler, B., Srebro, N., and Zhou, X.}
\newblock Semi-supervised learning with the graph laplacian: The limit of
  infinite unlabelled data.
\newblock {\em Advances in neural information processing systems 21\/} (2009).

\bibitem{snbc}
{\sc Nandanwar, S., and Murty, M.~N.}
\newblock Structural neighborhood based classification of nodes in a network.
\newblock In {\em Proceedings of the 22Nd ACM SIGKDD International Conference
  on Knowledge Discovery and Data Mining\/} (2016), KDD '16, pp.~1085--1094.

\bibitem{orlitsky2005estimating}
{\sc Orlitsky, A., et~al.}
\newblock Estimating and computing density based distance metrics.
\newblock In {\em Proceedings of the 22nd international conference on Machine
  learning\/} (2005), ACM, pp.~760--767.

\bibitem{deepwalk}
{\sc Perozzi, B., Al-Rfou, R., and Skiena, S.}
\newblock Deepwalk: Online learning of social representations.
\newblock In {\em Proceedings of the 20th ACM SIGKDD international conference
  on Knowledge discovery and data mining\/} (2014), ACM, pp.~701--710.

\bibitem{rifai2011manifold}
{\sc Rifai, S., Dauphin, Y., Vincent, P., Bengio, Y., and Muller, X.}
\newblock The manifold tangent classifier.
\newblock In {\em NIPS\/} (2011), vol.~271, p.~523.

\bibitem{dimple}
{\sc Rosenfeld, N., Nitzan, M., and Globerson, A.}
\newblock Discriminative learning of infection models.
\newblock In {\em Proceedings of the Ninth ACM International Conference on Web
  Search and Data Mining\/} (2016), WSDM '16, pp.~563--572.

\bibitem{flip}
{\sc Saha, T., Rangwala, H., and Domeniconi, C.}
\newblock Flip: active learning for relational network classification.
\newblock In {\em Joint European Conference on Machine Learning and Knowledge
  Discovery in Databases\/} (2014), Springer, pp.~1--18.

\bibitem{linqs}
{\sc Sen, P., Namata, G.~M., Bilgic, M., Getoor, L., Gallagher, B., and
  Eliassi-Rad, T.}
\newblock Collective classification in network data.
\newblock {\em AI Magazine 29}, 3 (2008), 93--106.

\bibitem{sindhwani2005beyond}
{\sc Sindhwani, V., Niyogi, P., and Belkin, M.}
\newblock Beyond the point cloud: from transductive to semi-supervised
  learning.
\newblock In {\em Proceedings of the 22nd international conference on Machine
  learning\/} (2005), ACM, pp.~824--831.

\bibitem{mad}
{\sc Talukdar, P.~P., and Crammer, K.}
\newblock New regularized algorithms for transductive learning.
\newblock In {\em Joint European Conference on Machine Learning and Knowledge
  Discovery in Databases\/} (2009), Springer, pp.~442--457.

\bibitem{vishnoi2013Lxb}
{\sc Vishnoi, N.~K., et~al.}
\newblock Lx= b.
\newblock {\em Foundations and Trends in Theoretical Computer Science 8}, 1--2
  (2013), 1--141.

\bibitem{vonLuxburg2010getting}
{\sc von Luxburg, U., Radl, A., and Hein, M.}
\newblock Getting lost in space: Large sample analysis of the resistance
  distance.
\newblock In {\em Advances in Neural Information Processing Systems\/} (2010),
  pp.~2622--2630.

\bibitem{planetoid}
{\sc Yang, Z., Cohen, W.~W., and Salakhutdinov, R.}
\newblock Revisiting semi-supervised learning with graph embeddings.
\newblock In {\em Proceedings of the 33rd International Conference on
  International Conference on Machine Learning\/} (2016), ICML'16, pp.~40--48.

\bibitem{llgc}
{\sc Zhou, D., Bousquet, O., Lal, T.~N., Weston, J., and Sch{\"o}lkopf, B.}
\newblock Learning with local and global consistency.
\newblock In {\em NIPS\/} (2003), vol.~16, pp.~321--328.

\bibitem{labelprop}
{\sc Zhu, X., Ghahramani, Z., Lafferty, J., et~al.}
\newblock Semi-supervised learning using gaussian fields and harmonic
  functions.
\newblock In {\em ICML\/} (2003), vol.~3, pp.~912--919.

\end{thebibliography}


\begin{thebibliography}{1}

\bibitem{cohen2014sketch}
{\sc Cohen, E., Delling, D., Pajor, T., and Werneck, R.~F.}
\newblock Sketch-based influence maximization and computation: Scaling up with
  guarantees.
\newblock In {\em Proceedings of the 23rd ACM International Conference on
  Conference on Information and Knowledge Management\/} (2014), ACM,
  pp.~629--638.

\bibitem{gadde2014active}
{\sc Gadde, A., Anis, A., and Ortega, A.}
\newblock Active semi-supervised learning using sampling theory for graph
  signals.
\newblock In {\em Proceedings of the 20th ACM SIGKDD international conference
  on Knowledge discovery and data mining\/} (2014), ACM, pp.~492--501.

\bibitem{mgr2016influence}
{\sc Gomez-Rodriguez, M., Song, L., Du, N., Zha, H., and Sch\"{o}lkopf, B.}
\newblock Influence estimation and maximization in continuous-time diffusion
  networks.
\newblock {\em ACM Trans. Inf. Syst. 34}, 2 (Feb. 2016), 9:1--9:33.

\bibitem{gu2012towards}
{\sc Gu, Q., and Han, J.}
\newblock Towards active learning on graphs: An error bound minimization
  approach.
\newblock In {\em Data Mining (ICDM), 2012 IEEE 12th International Conference
  on\/} (2012), IEEE, pp.~882--887.

\bibitem{guillory2012active}
{\sc Guillory, A., and Bilmes, J.~A.}
\newblock Active semi-supervised learning using submodular functions.
\newblock In {\em {UAI} 2011, Proceedings of the Twenty-Seventh Conference on
  Uncertainty in Artificial Intelligence\/} (2011), pp.~274--282.

\bibitem{kempe2003maximizing}
{\sc Kempe, D., Kleinberg, J., and Tardos, {\'E}.}
\newblock Maximizing the spread of influence through a social network.
\newblock In {\em Proceedings of the ninth ACM SIGKDD international conference
  on Knowledge discovery and data mining\/} (2003), ACM, pp.~137--146.

\bibitem{snbc}
{\sc Nandanwar, S., and Murty, M.~N.}
\newblock Structural neighborhood based classification of nodes in a network.
\newblock In {\em Proceedings of the 22Nd ACM SIGKDD International Conference
  on Knowledge Discovery and Data Mining\/} (2016), KDD '16, pp.~1085--1094.

\bibitem{flip}
{\sc Saha, T., Rangwala, H., and Domeniconi, C.}
\newblock Flip: active learning for relational network classification.
\newblock In {\em Joint European Conference on Machine Learning and Knowledge
  Discovery in Databases\/} (2014), Springer, pp.~1--18.

\bibitem{linqs}
{\sc Sen, P., Namata, G.~M., Bilgic, M., Getoor, L., Gallagher, B., and
  Eliassi-Rad, T.}
\newblock Collective classification in network data.
\newblock {\em AI Magazine 29}, 3 (2008), 93--106.

\end{thebibliography}

\documentclass[semisupinf_aistats]{subfiles}
\clearpage
\part*{Supplamentary Material}
\setcounter{section}{0}

\section{Proof of Proposition 2 in Main Text}
In this section we prove the correctness of our algorithm 2.
The proof considers the more general CTIC infection dynamics
and allows for node features or priors (via a penalty function).

%
In the infection dynamics presented in the paper, 
once a node's label is set, it remains fixed.
In contrast, during the course of the algorithm's run,
a node's label may change with each distance update.
It therefore remains to show that the algorithm
outputs the desired labels.
For basing our claim it will be easier to assume that instead
of initially inserting all seed nodes into $Q$,
we add a dummy root node $\dummy$ to $V$,
with edges of length $w_{\dummy s}=0$ to all $s \in \seed$,
and initialize $Q$ to include only $\dummy$.
It is easy to see that after extracting $\dummy$ from $Q$,
we return to our original algorithm.

Recall that the standard single-source Dijkstra algorithm
offers three important guarantees:
(1) the estimated distances of extracted nodes is correct
(and remains unchanged),
(2) nodes are extracted in increasing order of their true distance, and
(3) the distance estimates always upper-bound
the true distances.

Let $v \in \unlbld$ be a node that has just been extracted,
and assume by induction that the labels of all
previously extracted nodes (which include all seed nodes)
are correct.\footnote{
	Correct in the sense of the algorithm,
	not in the sense of their true labels.}
The above guarantees tell us that the distance from $\dummy$
to $v$ is correct, and all nodes on the shortest path from
$\dummy$ to $v$ have already been extracted.
This is true even when a penalty is incurred,
as it can only increase the distance estimate.
As these nodes are assumed to be correctly labeled,
$v$ inherits the correct label as well,
as by construction its shortest path from
$\dummy$ goes through exactly one seed node.
The correctness of the labels of the seed nodes gives the
induction basis, which concludes the proof.

\section{Extensions} \label{sec:extensions}
In this section we give an in-depth description of several useful extensions of our method
that were briefly discussed in the main text.
These include applying out method to the Linear Threshold model,
incorporating node features and priors into the infection dynamics,
and a framework for using our method in an active SSL setting.

\subsection{The Linear Threshold model}
In this section we show how InfProp can be applied with the
Linear Threshold (LT) dynamics,
rather than the IC or CTIC dynamics discussed in the main text.
This includes adapting the algorithm for computing expected labels
to the LT model, as well as supporting node features and priors.

The input to the LT model is a weighted graph $G=(V,E,W)$ and an initial
set of infected seed nodes $\seed$.
We assume that weights are positive, and that for each $v \in V$, the sum of incoming weights
$\sum_v W_{uv}$ is at most 1 (though it can be strictly less than 1).
Before the process begins, each node $u$ is assigned
a threshold $\eta_u$ sampled uniformly at random from the interval $[0,1]$.
The dynamics then progress in discrete time steps,
where at time $t$, a susceptible node $v$ becomes infected 
if the weighted sum of its infected neighbors exceeds its threshold.
Denoting by $I_u(t)$ an indicator of whether $u$ is infected at time $t$,
$v$ is infected at time $t$ if:
\begin{equation}
\sum_{u} W_{uv} I_v(t-1) \ge \eta_u  
\label{eq:lt_update}
\end{equation}
Note that the randomness in this model comes from the threshold $\eta$;
given $\eta$, the dynamics are deterministic.

The authors of \cite{kempe2003maximizing} show that the LT model
can also be equivalently expressed via a graphical perspective
using active edge sets.
Here, however, edges are no longer sampled independently.
Instead, for each node $v$, only at most one incoming edge will become active
in each instance.
Specifically, for each node $v$, each incoming edge $(u,v) \in E$ is selected
to be the (only) edge with probability $p_{uv} = W_{uv}$,
and with probability $W_{-u} = 1-\sum_v W_{uv}$ no incoming edges are activated.
Then, for a given instance, $v$ is infected if and only if
there is an active path to $v$ from some seed node in $\seed$.

An interesting interpretation of the above is that the chosen active edge $(u,v)$
can be thought of as corresponding to the node $u$ whose infection
triggered the infection of $v$ by crossing the threshold.
Under this view, a label-dependent specification of the above model
is one where $v$ inherits its label from its triggering neighbor $v$,
which we refer to as his infector.
This allows the model to be applied to the competitive setting which we consider.
In terms of implementation, the only necessary modification to the algorithm
is the way in which active edges are sampled.

The competitive LT model can also incorporate node priors using penalty terms.
Specifically, the node-label prior $\prior_{v\ell}$ 
will induce a multiplicative penalty $\pnlty_{v}(\ell) \in [0,1]$ on the
original weights $W_{uv}$ when $u$ tries to infect $v$ with label $\ell$.
Thus, given that $u$ has label $\ell$, the penalty reduces the probability
that it will be the infector of $v$.
To implement this, when $u$ is expanded, the edge $(u,v)$
is sampled to be active with the penalized probability,
and all other incoming weights (including the complementing $W_{-u}$)
are re-normalized.


\subsection{Incorporating node features and priors} \label{sec:priors}
In addition to the graph,
many network-based datasets include node or edge features.
These can be used to generate node-specific class priors.
In this section we describe a novel generalization of the competitive infection models introduced above
which incorporates class priors into the dynamics.
In this setting, our approach is to first train
a probabilistic classifier (e.g., logistic regression)
on the labeled seed set,
and then use its predictions on the unlabeled nodes
as a prior for our model.

Our method utilizes node priors by transforming
them into penalties on incubation times.
Consider a single instance of an infection process.
Assume node $u$ has just been infected
with label $\ell \in \Y$, and succeeded in its 
attempt to infect node $v$ with an incubation time of $\delta_{uv}$.
If $\delta_{uv}$ is small, then it is very likely that $v$
will get infected with $\ell$ as well.
On the other hand, if $\delta_{uv}$ is large,
then other nodes might have a chance to infect $v$ with
other labels.
This motivates the idea of further \emph{penalizing}
the infection time of a node according to its prior.
We do this by adding a label-dependent
penalty $\pnlty_{v}(\ell)$ to
$\delta_{uv}$,
as a function of the prior $\prior_{v\ell}$.
We use the link function
$\pnlty_{v}(\ell) = -\log(\prior_{v\ell})$,
which maps low priors into large penalties,
and high priors into low penalties,
where $\prior_{v\ell}=1$ entails no penalty.
Hence, setting $\prior_{v\ell}=1$ for all $v,\ell$
recovers the original model.

Note that while the priors are deployed locally,
their effect is in fact global, as penalizing a node's
infection time delays the potential propagation
of its acquired label throughout the graph.
This increases the significance of nodes which
are central to the infection process,
and reduces the significance of those which play a small
role in it,
a property captured by our notion of confidence.
The strength of the above formulation lies in its
ability to introduce non-linear label dependencies
to the actual infection dynamics.
To see this, we can write the original predictions as:
\begin{equation}
\f = 
\expect{\delta \sim \dist}{\Ancestor^\delta \mathcal{\seed}} =
\expect{\delta \sim \dist}{\Ancestor^\delta} \mathcal{\seed}
\end{equation}
where $\Ancestor^\delta_{v s} = \1{s=\ancestor(v)}$
indicates ancestors in $G^\delta$,
and $\mathcal{\seed}_{s \ell} = \1{y_s=\ell}$
indicates the seed nodes' true labels.
This shows that predictions are non-linear in the
propagation of the seed nodes, but linear in the labels.
In the prior-dependent model,
the above no longer holds,
as activation times are now label-dependent.

\subsection{Confidence and Active Learning} \label{sec:conf_and_active}
Recall that a node $v$ has a probability $f_{v0}$ of not being infected by {\em any} label. 
This suggests a very natural measure of \emph{confidence} in our prediction, namely:
\begin{equation}
\conf_v(\seed) = 1 - \f_{v0} =  \sum_{\ell=1}^\nclasses \f_{v\ell}, \qquad
\conf(\seed) = \sum_v \conf_v(S)
\end{equation}
The function $\conf$ quantifies the confidence \emph{in the labeling}.
This is conceptually different
from confidence \emph{in a label}.
Our model supports both concepts distinctly.
The former is controlled by the activations $p$,
as they determine reachability in the active graph and are 
agnostic to labels.
The latter is controlled by $\theta$,
as it affects the speed of propagation of the labels.
%
%
%

The notion of confidence allows us to apply our method
to an active SSL setting.
Instead of assuming the seed is given as input,
in this setting we are allowed to \emph{choose}
the seed set, often under a cardinality constraint.
The goal is then to choose the seed set which leads to a good labeling.
Various graph-based 
notions have been suggested as objectives
for active seed selection, such as those based on
graph cuts \cite{guillory2012active},
graph signals \cite{gadde2014active},
and generalization error \cite{gu2012towards}.
Such methods however either optimize an
adversarial objective,
or simply offer a heuristic solution.
In contrast, using $\conf$ as a seed-selection criterion
offers an optimistic alternative,
as summing over all classes makes it indifferent
to the actual (latent) labels.

The confidence term $\conf$ coincides with the well-studied
notion of \emph{influence}, defined as the expected number
of nodes a seed will infect.
In \cite{kempe2003maximizing} it is shown that for various settings,
influence is submodular, and therefore admits to a
greedy $(1-1/\epsilon)$-approximation scheme.
Any algorithm for maximizing influence efficiently
(e.g., \cite{cohen2014sketch,mgr2016influence}),
can therefore be adopted for out setting.


\section{Non-Homogeneous Laplacian}

Here we prove that:
\[
\lap(\seed)_\ip \f = \Bias(\seed)
\]
where:
\[
\Bias_{u\ell}(\seed) = \sum_v b_{vu\ell}^{(\seed)},
\qquad
b_{vu\ell}^{(\seed)} = \covargs{}{\parmat_{vu}(\seed),Y_{u\ell}}
\]
For clarity we drop the notational dependence on $\seed$.
We begin by expanding $\f_{u\ell}$ using $Y$ and $T$:
\begin{align*}
\f_{u\ell} &= \expect{}{Y_{u\ell}} = \expect{}{T_{u \cdotp} Y_{\cdotp \ell}} \\
&= \expect{}{\sum_v T_{uv} Y_{v\ell}} \\
&= \sum_v \expect{}{T_{uv} Y_{v\ell}} \\
&= \sum_v \left(
 \expect{}{T_{uv}}\expect{}{Y_{v\ell}} + \covargs{}{T_{uv}Y_{v\ell}} \right) \\
 &= \sum_v \left(
 \expparmat_{uv} \f_{v\ell} + \bias_{uv\ell} \right)
\end{align*}
where the final step is true for the product of general random variables.
Rewriting in matrix form gives:
$
\f = \expparmat \f + \Bias
$. 
Rearranging we get:
$
(I - \expparmat) \f  = \lap \f = \Bias
$, as required.

The objective function can then be expressed as:
\begin{align*}
\| \lap f' - \Bias \|_2^2 &= 
\sum_u \sum_\ell \left( \lap_{u \cdotp} \f_{\cdotp \ell} - \Bias_{u\ell} \right)^2 \\
&= \sum_u \sum_\ell \left( \sum_v \lap_{uv} \f_{v\ell} - \bias_{uv\ell} \right)^2 \\
&= \sum_u \sum_\ell \left( \sum_v (\1{u=v} - \expparmat_{uv}) \f_{v\ell} - \bias_{uv\ell} \right)^2 \\
&= \sum_u \sum_\ell \left( f_{u\ell} - \sum_v (w_{uv} \f_{v\ell} + \bias_{uv\ell}) \right)^2 
\end{align*}
where $w_{uv} = \expparmat_{uv}$.
\section{Details for the Illustrative Synthetic Experiment}
Our general hypothesis in this work is that infection dynamics
are a good candidate for propagating label information over real networks.
To illustrate this, we designed a synthetic experimental setup
in which our goal was to capture the structure of real world networks.
One well-known property of such networks is that they
often have a community-like structure,
with many intra-community edges, but few inter-community edges. 
In many cases, only a few specific nodes within a community are also connected
to other communities. 
Hence, we randomly created small networks with the above properties.

Specifically, each network was set to have 3 (possibly overlapping) communities,
each with 64 nodes.
Nodes were randomly assigned into one community, and in each community,
8 nodes were randomly assigned to an additional community.
To account for some noise, all other edges were added with probability 0.05.
The seed set included one randomly chosen node from each community, giving $|\seed| = 3$.
The figure in the main text 
displays a random instance of the above setting,
providing both the instance specific accuracies,
as well as the average accuracy over 1,000 random instances.

Recall that InfProp can be interpreted both as the expected result
of a dynamic infection process and as a stochastic ensemble of shortest paths.
We therefore compared our method to two baselines.
To compare the dynamics, we used Label Propagation (LabelProp)
which is based on the more standard random-walk dynamics.
As we argue in the text, these dynamics are prone to getting stuck in
dense clusters.
As can be seen, while InfProp provides almost exact predictions,
the predictive values of LabelProp are almost uniform and hence extremely error-prone.
This demonstrates the inability of label information to propagate efficiently
over the network.

To demonstrate the power of using a stochastic ensemble of paths,
we compared to simply setting labels according to the deterministic shortest
paths given by the original graph.
While correctly classifying most labels, shortest paths can be very sensitive
to cross-community or noisy edges.
In contrast, InfProp mitigates this noise
by considering a distribution over shortest-paths.


\section{Datasets}
We evaluated our method on various learning tasks
over three collections of benchmark datasets, which include network based datasets for
multi-class learning with features
\cite{linqs},
multi-class learning without features
\cite{flip}, 
and multi-label learning
\cite{snbc}. 
The following table provides some statistics.


\begin{table}[!h]
  \centering
	\fontsize{7pt}{10.5pt}\selectfont
	\setlength{\tabcolsep}{4pt}
  \begin{tabular}{|c|lrrccc|}
\cline{2-7}    \multicolumn{1}{r|}{} & Dataset & \multicolumn{1}{c}{Nodes} & \multicolumn{1}{c}{Edges} & Classes   & Features   & Avg. $|y|$ \bigstrut\\
    \hline
    \multirow{5}[2]{*}{\begin{sideways}Multiclass\footnotemark[\getrefnumber{foot:flip}]\end{sideways}} & CoRA  &              2,708  &                 5,278  & 7     & - & 1 \bigstrut[t]\\
          & DBLP  &              5,329  &              21,880  & 6     & -     & 1 \\
          & Flickr &              7,971  &            478,980  & 7     & -     & 1 \\
          & IMDb  &              2,411  &              12,255  & 22    & -     & 1 \\
          & Industry &              2,189  &              11,666  & 12    & -     & 1 \bigstrut[b]\\
    \hline
    \multirow{3}[2]{*}{\begin{sideways}Features\footnotemark[\getrefnumber{foot:linqs}]\end{sideways}} & CiteSeer &              3,132  &                 4,713  & 6     &         3,703  & 1 \bigstrut[t]\\
          & CoRA  &              2,708  &                 5,278  & 7     &         1,433  & 1 \\
          & PubMed &            19,717  &              44,324  & 3     &            500  & 1 \bigstrut[b]\\
    \hline
    \multirow{6}[2]{*}{\begin{sideways}Multilabel\footnotemark[\getrefnumber{foot:snbc}]\end{sideways}} & Amazon &            83,742  &            190,097  & 30    & -     & 1.546 \bigstrut[t]\\
          & CoRA  &            24,519  &              92,207  & 10    & -     & 1.004 \\
          & IMDb  &            19,359  &            362,079  & 21    & -     & 2.300 \\
          & PubMed &            19,717  &              44,324  & 3     & -     & 1 \\
          & Wikipedia &            35,633  &              49,538  & 16    & -     & 1.312 \\
          & YouTube &            22,693  &              96,361  & 47    & -     & 1.707 \bigstrut[b]\\
    \hline
    \end{tabular}%

  \label{tbl:data}%
\end{table}%




\end{document}